\documentclass[runningheads]{llncs}
\usepackage{times}
\usepackage{latexsym}
\usepackage{multirow}
\usepackage{graphicx}
\usepackage{pdfpages}
\usepackage{pgfplots}
\pgfplotsset{compat=1.18}

\usepackage{fancyvrb,new verbs,xcolor}

\definecolor{cverbbg}{gray}{0.93}

\newenvironment{lcverbatim}
 {\SaveVerbatim{cverb}}
 {\endSaveVerbatim
  \flushleft\fboxrule=0pt\fboxsep=.5em
  \colorbox{cverbbg}{%
    \makebox[\dimexpr\linewidth-2\fboxsep][l]{\BUseVerbatim{cverb}}%
  }
  \endflushleft
}

\usepackage[T1]{fontenc}
\usepackage[utf8]{inputenc}

\usepackage{microtype}

\usepackage{booktabs}
\usepackage{xcolor}
\usepackage{pifont}
\usepackage{amsmath,amssymb,amsfonts}%
\usepackage{textcomp}%
\usepackage{manyfoot}%
\usepackage{listings}%
\usepackage{subcaption}
\usepackage{caption}
\usepackage{pgf-pie}  
\usepackage{url}

\definecolor{roBlue}{rgb}{0,0.169,0.6}
\definecolor{roYellow}{rgb}{0.988,0.82,0.086}
\definecolor{roRed}{rgb}{0.808,0.067,0.149}

\definecolor{myRed}{rgb}{0.808,0.067,0.149}
\definecolor{myGreen}{rgb}{0.067,0.708,0.149}

\usepackage[T1]{fontenc}
\usepackage{graphicx}
\begin{document}
\title{\textcolor{roBlue}{Po}\textcolor{roYellow}{Pre}\textcolor{roRed}{Ro}: A New Dataset for \textcolor{roBlue}{Po}pularity \textcolor{roYellow}{Pre}diction of \textcolor{roRed}{Ro}manian Reddit Posts}
\titlerunning{PoPreRo: A New Dataset for Popularity Prediction of Romanian Reddit Posts}
% If the paper title is too long for the running head, you can set
% an abbreviated paper title here
%
\author{Ana-Cristina Rogoz \and Maria Ilinca Nechita \and Radu Tudor Ionescu\thanks{Corresponding author: \email{raducu.ionescu@gmail.com}}}
\institute{Department of Computer Science\\
University of Bucharest\\
14 Academiei, Bucharest, Romania}
\authorrunning{A.C. Rogoz et al.}

% First names are abbreviated in the running head.
% If there are more than two authors, 'et al.' is used.
%

%
\maketitle              % typeset the header of the contribution
\begin{abstract}
We introduce PoPreRo, the first dataset for \textbf{Po}pularity 
\textbf{Pre}diction of \textbf{Ro}manian posts collected from Reddit. The PoPreRo dataset includes a varied compilation of post samples from five distinct subreddits of Romania, totaling 28,107 data samples. Along with our novel dataset, we introduce a set of competitive models to be used as baselines for future research. Interestingly, the top-scoring model achieves an accuracy of $61.35\%$ and a macro $F_1$ score of $60.60\%$ on the test set, indicating that the popularity prediction task on PoPreRo is very challenging. Further investigations based on few-shot prompting the Falcon-7B Large Language Model also point in the same direction. We thus believe that PoPreRo is a valuable resource that can be used to evaluate models on predicting the popularity of social media posts in Romanian. We release our dataset at {\small{\url{https://github.com/ana-rogoz/PoPreRo}}}.
\keywords{natural language processing  \and reddit popularity \and popularity detection \and virality detection  \and Romanian \and LLM prompting.}
\end{abstract}
\section{Introduction}

Understanding the factors influencing the popularity of social media posts represents a critical and multifaceted challenge for NLP research. Social media platforms generate vast amounts of user-created content, offering a unique window into real-time public discourse and collective attention. Analyzing what resonates with audiences goes beyond just sentiment analysis, demanding nuanced NLP techniques to capture humor, sarcasm, and the subtle cues that drive engagement. This pursuit fosters not only theoretical advancements but also practical applications across diverse fields, from marketing and public health to combating misinformation and predicting cultural trends. Studying social media popularity, therefore, is not just an interesting NLP problem, but a key to unlocking the true potential of language in the digital age.

% for which social media platforms it has been previously studied
So far, the phenomenon has been studied both for individual social media platforms, such as Instagram \cite{info11090453,zhang2019instagram,Purba2021InstagramPP,8066548}, Reddit \cite{Barnes_2021,kim2021predicting}, Twitter \cite{inproceedings-twitter,10.1145/2783258.2783401,Ma2013OnPT}, either as a whole phenomenon, for detecting popularity \cite{POECZE2018660,6650118}, or for generating engaging content \cite{FANG20241}.

% Focus on reddit.
Reddit, in particular, has been one of the most studied platforms in the ever-evolving landscape of online content. From gauging public opinion and identifying emerging trends to optimizing content recommendation systems and combating misinformation, accurate popularity detection offers a multitude of applications across various domains. There are existing datasets generated from Reddit content, studying several topics, from political conflicts \cite{zhu2022reddit}, to personality traits \cite{gjurkovic-snajder-2018-reddit}, language biases \cite{ferrer2020discovering,hada2022ruddit}, and mental health related topics, such as stress analysis \cite{turcan2019dreaddit}, depression \cite{8681445} and anxiety \cite{shen-rudzicz-2017-detecting}.

% lack of resources for romanian 
While existing Reddit datasets have played a crucial role in advancing NLP research, they predominantly focus on high-resource languages, such as English. This creates a bias towards high-resource languages in NLP models, neglecting the necessity of exploring NLP capabilities on less studied languages, such as Romanian.
% we introduce the first to our knowledge dataset 

We emphasize that what constitutes a popular (viral) post can vary across countries and regions, since the topics of interest can naturally change from one local community to another. This is because people are usually more influenced by major local events, e.g.~the war in Ukraine is still a major subject of discussion in Romania, a neighboring country of Ukraine, while the subject may have faded out in countries from other continents. This justifies the need to study the popularity prediction task across multiple countries, and consequently, in various languages. To this end, we introduce PoPreRo, the first dataset for \textbf{Po}pularity \textbf{Pre}diction of \textbf{Ro}manian posts collected from Reddit. We leverage this novel resource to explore popularity detection in a low-resource language, Romanian, establishing six diverse baselines for future comparative analysis.

\section{Dataset}
\subsection{Data Collection}

PoPreRo gathers Reddit posts from five different Romanian subreddit channels, which represent either one of the biggest cities in Romania or the country-wide subreddit. 
The subreddits are: Romania, Bucure\c{s}ti, Cluj, Ia\c{s}i and Timi\c{s}oara. These subreddits were collected at first using Reddit API, divided into JSON files  to extract the information needed for analyzing the popularity of each reddit post, such as title, content, number of comments, number of up and down votes. However, Reddit API has a limitation of 1000 requests for extraction of different data. Due to the large number of samples that we target for the dataset, the API could not provide all necessary data. Therefore, we use an open-source archive, from where the samples are collected. As mentioned above, all the data is stored in separate JSON files for each subreddit, containing relevant information for determining the popularity of posts. 

\subsection{Dataset Statistics} 

\begin{table}[!t]
\caption{Number of samples (\#posts) and number of tokens (\#tokens) for each subset in PoPreRo.}
\label{tab_PoPreRo}
\setlength\tabcolsep{4.2pt}
\begin{center}
%\footnotesize
\begin{tabular}{|l|r|r|r|r|r|r|}
\hline
\multirow{2}{*}{Set} 						& \multicolumn{2}{|c|}{Unpopular}     & \multicolumn{2}{|c|}{Popular}   & \multicolumn{2}{|c|}{Total}\\
\cline{2-7}
     						& \#posts		& \#tokens	        & \#posts		& \#tokens          & \#posts		& \#tokens\\
\hline
\hline
Training				    & 12,053		& 398,219         & 11,592 		& 560,580         & 23,645        & 958,799\\
Validation					& 1,059		& 75,742         & 1,054 		& 80,297         & 2,113       & 156,039\\
Test					    & 1,177		& 72,819        & 1,172 		& 93,268         & 2,349        & 168,867\\
\hline
Total						& 14,289		& 546,780         & 13,818  		& 734,145         & 28,107         & 1,283,705\\
\hline
\end{tabular}
\end{center}
\end{table}

The dataset comprises 28,107 samples (14,289 unpopular and 13,818 popular) containing over 1 million tokens in total (see detailed statistics in Table \ref{tab_PoPreRo}). 
Each sample consists of a title, a content, and a binary label, where the title and content are concatenated into a single text. We divide the posts into ``popular'' or ``unpopular'' based on the sum of upvotes and downvotes for each post, where the threshold between the two categories is given by the median number of votes (15). To enable consistent evaluation and comparison with future studies, we provide an official split with distinct training, validation, and test sets. Inspired by McHardy et al.~\cite{McHardy-NAACL-2019}, we utilize disjoint subreddits for each set, ensuring models cannot capitalize on knowledge of specific topics. To further mitigate potential biases arising from uneven topic or time distributions, we select posts from the same time frame across all subreddits. % during the crawling process.

\begin{figure}[h!]
\centering
\begin{tikzpicture}
\begin{axis}[
    ybar,
    bar width=0.40,
    height = 8cm,
    enlarge x limits=0.12,
    width=1.0\textwidth, 
    ymin=0,
    legend style={at={(0.5,-0.2)},
      anchor=north,legend columns=-1},
    ylabel={\#posts},
    xticklabels={[0-4),[4-8),[8-12),[12-16),[16-20),[20-24)},
    xtick={0,1,2,3,4,5},
    xlabel={time intervals},
    nodes near coords,
    nodes near coords align={vertical},
    ]
    
% Summing the values from train, validation, and test
\addplot coordinates {(0,816+78+57) (1,260+38+24) (2,2200+255+241) (3,3451+284+319) (4,2797+228+287) (5,2272+172+244)};
\addplot coordinates {(0,1050+87+67) (1,254+32+32) (2,1779+174+259) (3,3280+273+325) (4,3014+232+274) (5,2472+260+220)};
\legend{Popular, Unpopular}
\end{axis}
\end{tikzpicture}
\caption{Number of samples (\#posts) for each label (popular/unpopular), distributed by the time of posting. The 24 hours in a day are divided into six four-hour intervals. Best viewed in color.}
\label{fig:combined_bar_chart}
\end{figure}
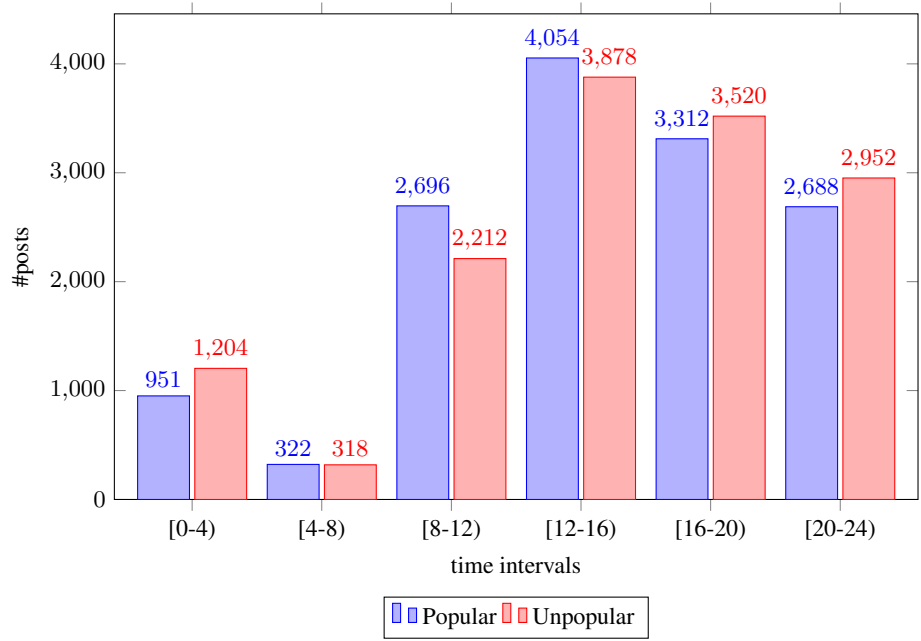

Additionally, to control for a potential bias related to the time of day when posts were submitted, we performed an analysis of post popularity by hour (see Table \ref{tab_hours}). We divided each day into four-hour intervals and categorized the number of popular and unpopular posts within each interval. The detailed results are presented in Figure \ref{fig:combined_bar_chart}. Notably, we observe a consistent trend across all time intervals for both popular and unpopular posts. This finding suggests that the hour of submission does not exert a significant influence on post popularity within our dataset.

\begin{table*}[t!]
\caption{Number of samples (\#posts) for each label (popular/unpopular), distributed by the time of posting for each subset in PoPreRo.}
\label{tab_hours}
\setlength\tabcolsep{4.2pt}
\begin{center}
%\footnotesize
\begin{tabular}{|l|l|r|r|r|r|r|r|}
\hline
\multirow{2}{*}{Set} 	& \multirow{2}{*}{Label}					& \multicolumn{6}{c|}{\#posts in time window (h)}\\
\cline{3-8}
     &						& [0-4) & [4-8) & [8-12) & [12-16) &  [16-20) & [20-24) \\
\hline
\hline
\multirow{2}{*}{Training} &		popular		    & 816 & 260 & 2,200 & 3,451 & 2,797 & 2,272 \\
\cline{2-8}
&		unpopular		    & 1,050 & 254 & 1,779 & 3,280 & 3,014 & 2,472\\
\hline
\multirow{2}{*}{Validation} &	popular				& 78 &  38 & 255 & 284 & 228 & 172 	\\
\cline{2-8}
 &	unpopular				& 87 & 32 & 174 & 273 & 232 & 260 \\
\hline
\multirow{2}{*}{Test}		&	popular		    & 57 & 24 & 241 & 319 & 287 & 244 \\
\cline{2-8}
		&	unpopular		    & 67 & 32 & 259 & 325 &274 & 220 \\
\hline
\end{tabular}
\end{center}
\end{table*}

\subsection{Preprocessing}

After gathering the data from Reddit, we implement a two-step preprocessing pipeline to ensure data quality and consistency. First, language identification was performed on post titles using FastText \cite{joulin2016bag} to filter out non-Romanian posts (filtered posts are not counted in Table \ref{tab_PoPreRo}). This step guarantees the linguistic homogeneity of the dataset. Subsequently, upvote/downvote scores are normalized to the $[0, 1]$ interval. Finally, a binary popularity label is assigned with respect to the median value of the normalized scores, which corresponds to 15 votes. This approach provides a clear threshold for distinguishing popular and unpopular posts. Notably, our data collection and labeling procedure is directly transferable to other languages.

\section{Methods}

To comprehensively evaluate the performance for the popularity prediction task on the newly introduced dataset, we establish six baseline approaches. Two of these baselines leverage state-of-the-art deep learning models for language processing. Another three baselines utilize various classifiers based on shallow or deep (frozen) features. Our final baseline uses a Large Language Model (LLM) based on in-context learning, also known as few-shot prompting. For all models, we use the concatenated title and content of each post as the input data.

\subsection{Fine-Tuned Ro-GPT2} 

Our first baseline relies on fine-tuning a Ro-GPT2 model \cite{niculescu2021rogpt2}, a large language model specifically trained on Romanian text. It is based on the original GPT2 architecture, but trained on a Romanian dataset consisting of over 1 million tokens. This allows it to capture the nuances and specificities of the Romanian language, making it more suitable for tasks involving Romanian than the general-purpose GPT2.
The Ro-GPT2 encoder is utilized to encode each text sequence into a list of token IDs. Subsequently, the model processes these tokens, generating corresponding 768-dimensional embeddings. We then incorporate a global average pooling layer to capture a Continuous Bag-of-Words (CBOW) representation for each text sequence. This representation is fed into a Softmax output layer comprising two neurons, each predicting the probability of belonging to either the unpopular or popular category. To assign the final class label, we apply the \emph{argmax} function on the two predicted probabilities. The entire model is fine-tuned for 5 epochs on mini-batches of 32 samples. We employ the Adam optimizer with decoupled weight decay (AdamW) \cite{Loshchilov-ICLR-2019} with a learning rate of $5\cdot 10^{-7}$ and $\epsilon=5\cdot 10^{-7}$.

\subsection{Fine-Tuned Ro-BERT} As our second baseline, we employ a fine-tuned Romanian Bidirectional Encoder Representations from Transformers (Ro-BERT) model \cite{Dumitrescu-EMNLP-2020}. Sharing the same transformer-based architecture as the original BERT \cite{Devlin-NAACL-2019}, Ro-BERT has been demonstrated to outperform multilingual BERT on various tasks, as reported by Dumitrescu et al.~\cite{Dumitrescu-EMNLP-2020}. Consequently, we anticipate Ro-BERT to be a strong baseline for our Romanian corpus.

Similarly to the previous baseline, we use the Ro-BERT encoder to encode each text into a list of token IDs. We keep the same design as before, where the model generates 768-dimensional embeddings, followed by a global average pooling layer which is fed into a Softmax output layer with two neurons. To assign the final class label, we apply the \emph{argmax} function on the two predicted probabilities. The entire model is fine-tuned for 10 epochs on mini-batches of 32 samples. We employ the AdamW optimizer \cite{Loshchilov-ICLR-2019} with a learning rate of $2\cdot 10^{-7}$ and the default value for $\epsilon$.

\subsection{Ro-BERT Embeddings + Logistic Regression}

For our third classification approach, we leverage pre-trained Ro-BERT embeddings in conjunction with a Logistic Regression (LR) classifier. Consistent with the fine-tuned Ro-BERT baseline, we first tokenize all input samples from the three datasets. Subsequently, we utilize the Ro-BERT model to extract 768-dimensional vector representations for each sample. These representations, corresponding to the final hidden layer of Ro-BERT, are then fed into the LR model for classification.

\subsection{FastText + SVM} 
The first shallow classification approach is based on FastText embeddings \cite{bojanowski2016enriching} and a Support Vector Machines (SVM) classifier. After textual cleaning and tokenization using NLTK's word tokenizer, we fine-tune a FastText model on the training corpus. This model provides word embeddings for train, validation, and test sets. For each text sample, the word embeddings are averaged to produce a 300-dimensional feature vector, which is subsequently passed to the SVM. Finally, we train the SVM classifier using the linear kernel and the regularization hyperparameter $C$ set to $10$.

\subsection{TF-IDF + Random Forest}
Our second shallow classification approach is based on the Term Frequency-Inverse Document Frequency (TF-IDF) representation and a Random Forest (RF) classifier. As for the previous method, we initiate the process by cleaning and tokenizing the text using NLTK's word tokenizer. Subsequently, we employed a TF-IDF vectorizer to quantify the importance of words within the corpus, generating numerical features for each document. These features are then used to train a Random Forest classifier. 

\subsection{Few-Shot LLM Prompting}

To explore the feasibility of large language models (LLMs) for post popularity prediction in PoPreRo, we employ a prompt-based approach utilizing the 7-billion parameter Falcon LLM \cite{falcon40b} (Falcon-7B). Due to computational limitations, we prompt the LLM with contexts comprising two unpopular and two popular examples. Subsequently, we attach an individual test sample to each prompt and ask the LLM to predict the corresponding label. 
Below, we illustrate the structure of our prompt via a concrete example:
\begin{lcverbatim}
PROMPT (Original): 
Text: 'Nu vreau sa mai traiesc pe aceasta planeta !'
Label: 'Popular'.

Text: 'Unde pot verifica compoziția unui produs?. Să 
testez de exemplu dacă ingredientele unui produs sunt 
într-adevăr acelea. Sau dacă niște tablete de vitamine 
chiar conțin vitamine. În ce proporții? Sau câtă vitamina 
A conține un morcov - unde pot verifica asta? Ceva 
laboratoare?' Label: 'Unpopular'.

Text: 'Azi a venit mitropolitul ardealului la noi la 
liceu să ne convingă să facem religie. Primul lucru 
care mi-a venit în cap când am văzut ce mașină și-a
parcat în curtea instituții..' Label: 'Popular'.

Text: 'Daca intereseaza pe cineva, sa stiti ca e reddit 
si in romana' Label: 'Unpopular'.

Text: 'Am prins niste fulgere faine zilele trecute' 
Label: 
\end{lcverbatim}

\begin{lcverbatim}
PROMPT (Translated): 
Text: 'I don't want to live on this planet anymore!'
Label: 'Popular'.

Text: 'Where can I check the composition of a product?. 
To test for example whether the ingredients of a 
product are indeed those. Or if some vitamin tablets
\end{lcverbatim}
\begin{lcverbatim}
actually contain vitamins. In what proportions? Or how 
much vitamin A contains a carrot - where can I check 
this? Some laboratories?' Label: 'Unpopular'.

Text: 'Today the metropolitan of Transylvania came to us 
at high school to convince us to do religion. First thing
that came to mind when I saw what car he has
parked in the courtyard of institutions..' Label: 
'Popular'.

Text: 'If anyone is interested, there's reddit in 
Romanian' Label: 'Unpopular'.

Text: 'I caught some fine lightning the other day' 
Label: 
\end{lcverbatim}

\section{Experiments}

\subsection{Evaluation}

Our binary classification experiments focus on predicting the popularity of text within the PoPreRo dataset. Each text sample is categorized as either popular or unpopular. To evaluate the performance of our models, we employ several metrics. For each class, we calculate precision (proportion of true positives among the identified positives) and recall (proportion of true positives with respect to all positives). Additionally, we aggregate these scores using macro $F_1$ and micro $F_1$ (accuracy) measures. 

\subsection{Hyperparameter Tuning}

The hyperparameters of all models are determined via grid search. For the transformer-based methods (Ro-BERT, Ro-GPT2), we employ a grid search over the maximum number of input tokens in the set $\{50, 70, 100, 120, 150, 200 \}$, as well as the learning rate in the set $\{10^{-5}, 5\cdot 10^{-5}, 10^{-6}, 5\cdot 10^{-6}, 10^{-7}, 2\cdot 10^{-7}, 5\cdot 10^{-7}, 10^{-8}, 5\cdot 10^{-8}\}$ and the value of $\epsilon$ for AdamW in the set $\{10^{-6}, 10^{-7},10^{-8}\}$. 

For the FastText + SVM approach, we vary the FastText word-embeddings dimension ($\{150, 200, 300, 350\}$), the window size for the input ($\{2, 3, 4\}$), as well as the kernel (linear or RBF) and the parameter $C$ ($\{0.1, 1, 10, 100, 1000\}$) of the SVM classifier. Similarly, for the Ro-BERT + Logistic Regression approach, we run a search over the maximum numbers of Ro-BERT input tokens in the same set as before ($\{50, 70, 100, 120, 150, 200 \}$) and test different penalty term values (`l1', `l2', `elastic net' or `None') for the classifier.

Lastly, for the TF-IDF + Random Forest method, we vary the minimum ($\{4, 5, 6\}$) and maximum ($\{0.6, 0.7, 0.8\}$, in percentages) document frequency of the TF-IDF Vectorizer, together with the number of decision trees in the set $\{50, 100, 150, 200\}$ for the Random Forest classifier.

All other hyperparameters are set to their default values. Please note that we release the code to reproduce all baselines, along with the PoPreRo dataset\footnote{\url{https://github.com/ana-rogoz/PoPreRo}}.

\subsection{Results} 

We present the results of our five baselines on the PoPreRo validation and test sets in Table~\ref{tab_results_full}. We find that Ro-GPT2 exhibits the best performance, with an accuracy (micro $F_1$) and a macro $F_1$ score above $0.6$ on both validation and test sets, in contrast to the other baselines which seem to perform similarly well on the validation set, but reach worse performance on the test set. 

\begin{table}[!t]
\caption{Validation and test results of the six baselines. The random chance baseline is added as reference. There is no hyperparameter tuning for Falcon-7B LLM, so the model is directly applied on the test set (using in-context learning). The best score on each subset and for each metric is highlighted in bold.}
\label{tab_results_full}
\setlength\tabcolsep{4.2pt}
\begin{center}
%\footnotesize
\begin{tabular}{|c|l|c|c|c|c|c|c|}
\hline 
\multirow{2}{*}{Set} & \multirow{2}{*}{Method} & \multirow{2}{*}{Acc.} & Macro & \multicolumn{2}{|c|}{Unpopular} & \multicolumn{2}{|c|}{Popular} \\
\cline{5-8}
&  & & $F_1$ & Prec. & Rec. & Prec. & Rec.\\
\hline
\hline
% Majority class & 0.5011 & 0.3338 & 0.5011 & 1 & 0 & 0&  0.5010& 0.3337& 0.5010 & 1 & 0 & 0 \\
\multirow{6}{*}{\rotatebox[origin=c]{90}{Validation}} & Random chance & 0.4998 & 0.4999 & 0.4988 & 0.5011 & 0.5011 & 0.4988  \\
\cline{2-8}
 & Fine-tuned Ro-GPT2  & 0.6525	& 0.6397	& 0.6157	& 0.8097	& 0.7351	& 0.4986 \\
& Fine-tuned Ro-BERT & 0.6343	& 0.6278	& 0.6189	& 0.6995	& 0.6411	& \textbf{0.5562} \\
& FastText + SVM & 0.6677	& 0.6624	& 0.6348	& 0.7920	& 0.7225 & 0.5431 \\
& TF-IDF + RF & 0.6535	& 0.6395	& 0.6107	& 0.8497	& 0.7519	& 0.4568 \\
& Ro-BERT + LR & \textbf{0.6824}	& \textbf{0.6721}	& \textbf{0.6354} & \textbf{0.8582} & \textbf{0.7807} & 0.5061 \\
\hline \hline
\multirow{7}{*}{\rotatebox[origin=c]{90}{Test}} & Random chance & 0.4998 & 0.4999 & 0.5010 & 0.4989 & 0.4989 & 0.5010 \\
\cline{2-8}
 & Fine-tuned Ro-GPT2  & \textbf{0.6135}	& \textbf{0.6060} & \textbf{0.6146} &	0.6331 &	0.6145 & 0.5933 \\
& Fine-tuned Ro-BERT & 0.5605	& 0.5489 & 0.5505 &	0.6611 &	0.5767 &	0.4565 \\
& FastText + SVM & 0.5644	& 0.5637 & 0.5718 &	0.5208 &	0.5583 &	\textbf{0.6083} \\
& TF-IDF + RF &  0.5759	& 0.5729 & 0.5661 &	0.6584 &	0.5897 &	0.4931 \\
& Ro-BERT + LR &  0.5998 &  0.5973 & 0.5873 & 0.6771 & \textbf{0.6169} & 0.5221\\
\cline{2-8}
& Few-shot prompted Falcon-7B & 0.4143 & 0.4126& 0.4143 & \textbf{0.7904} &0.5537 & 0.1887 \\
\hline
\end{tabular}
\end{center}
\end{table}

Evaluating the two state-of-the-art transformer models, Ro-GPT2 and Ro-BERT, reveals some interesting findings. While both achieve comparable accuracy on the validation set ($0.6525$ for Ro-GPT2 and $0.6343$ for Ro-BERT), Ro-GPT2 clearly outperforms Ro-BERT on the test set, indicating the superior ability of the former model to generalize to unseen data. Analyzing the precision-recall trade-off, we observe a shared propensity for both models to exhibit higher recall for the ``popular'' category, followed by a shift towards higher precision when identifying the ``unpopular'' class.

\begin{table}[!t]
\caption{Examples of relevant terms for popular posts, learned by the fine-tuned Ro-BERT and SVM models.}
\label{tab_most_predictive_patterns_popular}
\vspace{0.2cm}
\setlength\tabcolsep{3.8pt}
\centering
\begin{tabular}{|c|p{0.15\linewidth}|p{0.32\linewidth}|p{0.34\linewidth}|}
\hline
Model & {Topic} &  Example & Translation \\
\hline 
\hline
%%%%% BERT
\multirow{9}{*}{\rotatebox[origin=c]{90}{Ro-BERT}} & Call to action & \textit{``pentru cei care vor s\u{a} se implice activ ''} & \textit{``for those who want to be actively involved''} \\
\cline{3-4}
       &             & \textit{``ar fi interesati de un voluntariat''} & \textit{``would be interested in volunteering''} \\
\cline{2-4}
& News & \textit{``\^{i}ncep s\u{a}p\u{a}turile la metrou''} & \textit{``excavations begin at the subway''}\\
\cline{3-4}
           &  &       \textit{``un nou residence la ``doar 20 de minute'' de Centru''}     & \textit{``a new residence building ``only 20 minutes'' from the center''}\\
           \cline{2-4}
           &  Events         & \textit{``Seara de film la Casa Tineretului''} & \textit{``Movie night at the Youth House''}\\
\hline
%%%%%% SVM
\multirow{5}{*}{\rotatebox[origin=c]{90}{SVM}} & News    & \textit{``mic protest la primaria capitalei``}   & \textit{``small protest at Bucharest City Hall``} \\
\cline{2-4}
 & Local transport  & \textit{``am vazut ca este tren de la gara de nord la aeroport aproape la fiecare ora``}  & \textit{``I saw that there is a train from Gara de Nord to the airport almost every hour``}\\
\hline
\end{tabular}
\end{table}

The FastText + SVM, TF-IDF + RF and Ro-BERT + LR models achieve comparable performance. All three models obtain accuracy rates higher than $65\%$ on the validation set, which drop below $60\%$ on the test set. In terms of precision and recall, almost all of them achieve higher precision for the ``popular'' category on both validation and test sets, with one exception being the FastText + SVM method on the test set, where the precision on the two classes is comparable. 
A distinctive behavior of the three models is that the TF-IDF + RF obtains a higher recall for the ``popular'' category, while FastText + SVM and Ro-BERT + LR attain a higher recall for the ``unpopular'' category.

% \begin{table}[!t]
% \setlength\tabcolsep{2.2pt}
% \begin{center}
% \footnotesize
% \begin{tabular}{|l|c|c|c|c|c|c|}
% \hline 
% \multirow{3}{*}{Method} & \multicolumn{6}{|c|}{Test} \\
% \cline{2-7}
%  						&  \multirow{2}{*}{Acc.} & Macro & \multicolumn{2}{|c|}{Unpopular} & \multicolumn{2}{|c|}{Popular} \\
% \cline{4-7} 
% & & $F_1$ & Prec. & Rec. & Prec. & Rec.\\
% \hline
% \hline
% % Majority class & 0.5011 & 0.3338 & 0.5011 & 1 & 0 & 0&  0.5010& 0.3337& 0.5010 & 1 & 0 & 0 \\
% Random chance & 0.4998 & 0.4999 & 0.5010 & 0.4989 & 0.4989 & 0.5010 \\
% \hline
% Few-shot LLM & 0.4143 & 0.4126& 0.4143 & 0.7904 &0.5537 & 0.1887\\
% \hline
% \end{tabular}
% \end{center}
% \caption{Test results of the few-shot prompted LLM against Random chance classifier.}
% \label{tab_results_llm}
% \end{table}

Table~\ref{tab_results_full} also shows the results on the test set of our few-shot prompted LLM. While this approach exhibits a bias similar to our other baselines, favoring recall for unpopular predictions and precision for popular ones, its overall performance falls below that of a random chance classifier. This suggests a limitation in the generalization capacity of LLMs to the popularity prediction task, particularly for languages with limited online resources, such as Romanian.

\subsection{Discriminative Feature Analysis} 

We analyze the discriminative features learned by the fine-tuned Ro-BERT and by the FastText + SVM. The motivation behind this analysis is to validate that the decisions of these models are not based on some biases that escaped our data collection, but on actual data understanding. 

For the Ro-BERT model, we use the Captum \cite{kokhlikyan2020captum} library via its Layer Integrated Gradients method to infer valuable insights from the fine-tuned model. This technique delves into the BERT embedding layer, attributing importance scores to individual input words which led to the final label prediction.

To find the words with higher influence on the decisions given by the SVM, we consider the cosine similarities between the primal weights of the SVM and the FastText embedding of each word. We sort the words based on the similarity values, and keep the first 10 and last 10 words from the sorted list as features for the positive (``popular'') and negative (``unpopular'') classes, respectively.

\begin{table}[!t]
\caption{Examples of relevant terms for unpopular posts, learned by the fine-tuned Ro-BERT and SVM models.}
\label{tab_most_predictive_patterns_unpopular}
\vspace{0.2cm}
\setlength\tabcolsep{3.8pt}
\centering
\begin{tabular}{|c|p{0.16\linewidth}|p{0.31\linewidth}|p{0.33\linewidth}|}
\hline
Model & {Topic} &  Example & Translation \\
\hline 
\hline
%%%%% BERT
\multirow{10}{*}{\rotatebox[origin=c]{90}{Ro-BERT}} & Proper names &  \textit{``Palatul Roznovanu''} & \textit{``Roznovanu palace''}\\
\cline{3-4}
& & \textit{``Ceau\c{s}escu''} & \textit{``Ceau\c{s}escu''} \\
\cline{3-4}
        &              & \textit{``\^{i}n Timi\c{s}oara''} & \textit{``in Timi\c{s}oara''} \\
       \cline{3-4}
\cline{2-4}
& Seeking advice  &  \textit{``terenuri ok de baschet \^{i}n...''} & \textit{``ok basketball courts in...''} \\
\cline{3-4}
&         & \textit{``print shop pentru poze mari \^{i}n ...''} & \textit{``print shop for big pictures in ...''} \\
\cline{2-4}
&  Mundane problems        & \textit{``Se \^{i}nchide circula\c{t}ia''} & \textit{``traffic is closed''}\\
\cline{3-4}
&  & \textit{``construim blocuri \^{i}ntre case''} & \textit{``building apartment building between houses''} \\
\hline
%%%%%% SVM
\multirow{5}{*}{\rotatebox[origin=c]{90}{SVM}} & City names    &  \textit{``bucuresti''}    & \textit{``bucharest''}\\
\cline{2-4}
 & Seeking advice   &\textit{``cunoasteti un loc de facut tatuaj temporar personalizat ''}   & \textit{``do you know a place to do custom temporary tattoo''} \\
\cline{2-4}
& Opinion sharing  &\textit{``lumea ca se plange de targul de craciun de anul acesta ''} &\textit{``people complain about this year's Christmas market''} \\
\hline
\end{tabular}
\end{table}

In Tables~\ref{tab_most_predictive_patterns_popular} and~\ref{tab_most_predictive_patterns_unpopular}, we present a few examples of interesting patterns that were picked up by the models. 
In predicting post popularity, the Ro-BERT model demonstrates a bias toward content reflecting current trends, including news and events, and posts encouraging community engagement through calls to action. Conversely, references to proper nouns like city names or historical landmarks appear to hinder popularity, as do posts seeking community advice or expressing dissatisfaction with platitudes. Similar to Ro-BERT, we find that the SVM labels posts that share news as popular, and posts by people seeking advice as unpopular.

\begin{table}[!t]
\caption{Examples of the most discriminative words for the popular and unpopular classes, selected according to the weights learned by the SVM model based on FastText features.}
\label{tab_most_popular_and_unpopular}
\setlength\tabcolsep{3.8pt}
\centering
\begin{tabular}{|p{0.1\linewidth}|p{0.15\linewidth}|p{0.15\linewidth}|}
\hline
 Label & Token & Weight \\
\hline
\hline 
\multirow{5}{*}{popular} & online & 5.974352  \\
\cline{2-3}
& dupa & 4.821379 \\
\cline{2-3}
& youtube & 4.121604 \\
\cline{2-3}
& asa & 4.08882 \\
\cline{2-3}
& cazul & 3.839789 \\
\hline
\hline
\multirow{5}{*}{unpopular} & toate & -4.089375 \\
\cline{2-3}
& un & -4.190036\\
\cline{2-3}
& google & -4.31336 \\
\cline{2-3}
& nia & -4.339616 \\
\cline{2-3}
& eu & -4.72841 \\
\hline
\end{tabular}
\end{table}

Furthermore, we extend the feature analysis for the SVM in order to determine the most discriminative words for the popular and unpopular classes. To achieve this, we determine the discriminative weight of each word based on the cosine similarity between the respective word embedding and the SVM weights. We sort the words according to their weights, and select the ones with the highest and lowest weights. In Table~\ref{tab_most_popular_and_unpopular}, we provide the five most discriminative words for the popular and unpopular classes, according to the SVM based on FastText features. We observe that posts mentioning ``online'' or ``youtube''  are more popular, likely because readers appreciate posts that provide links to YouTube videos. We also note the preference for posts that discuss particular cases/experiences, which are usually introduced by the word ``cazul'' (translated to ``case'' in English). On the other hand, posts that recommend searching on ``google'' are unpopular, as the readers consider such suggestions unhelpful. Moreover, discussing subjective perspectives, using the singular first person pronoun ``eu'', is again unpopular, likely because the readers appreciate more objective posts. 

\section{Conclusion}

In this paper, we introduced PoPreRo, the first publicly available dataset of Romanian Reddit posts dedicated to the task of popularity prediction. We collected 28,107 posts from five diverse Romanian subreddits, amounting to over 1 million tokens. Aiming to predict binary labels resulting from the sum of upvotes and downvotes for each post, we explored five distinct popularity detection methods and presented comparative results. We found that Ro-GPT2 significantly outperforms the other models.

Building upon our foundation, future research can further study popularity detection algorithms and delve deeper into the factors driving engagement on Romanian Reddit.

\section{Limitations}

It is crucial to acknowledge that Reddit's popularity in Romania might not be representative for the wider population. While Reddit offers a valuable platform for research due to its diverse communities and open discussions, its user base in Romania is comparatively smaller than other social media platforms, such as Facebook, Instagram, or YouTube. Furthermore, Reddit's API restricts data access, limiting historical data collection and imposing retrieval caps.

\section{Ethics Statement}

The data was collected from a publicly available Reddit archive, selecting five Romanian subreddits. The social media posts are freely accessible to the public without any type of subscription. As the data was collected from an archived public website (Reddit), we adhere to the European regulations\footnote{\url{https://eur-lex.europa.eu/eli/dir/2019/790/oj}} that allow researchers to use data in the public web domain for non-commercial research purposes. We thus release our corpus as open-source under a non-commercial share-alike license agreement, namely CC BY-NC-SA 4.0\footnote{\url{https://creativecommons.org/licenses/by-nc-sa/4.0/}}.

We acknowledge that some posts could refer to certain people, e.g.~public figures in Romania. Following GDPR regulations, we will remove all references to a person, upon receiving removal requests via an email to any of the authors.

%
% ---- Bibliography ----
%
% BibTeX users should specify bibliography style 'splncs04'.
% References will then be sorted and formatted in the correct style.
%
% \bibliographystyle{splncs04}
% \bibliography{mybibliography}
%
\bibliographystyle{splncs04}
\bibliography{anthology} 

\end{document}